\documentclass{article}

\usepackage[final]{nips_2017}

\usepackage[utf8]{inputenc} % allow utf-8 input
\usepackage[T1]{fontenc}    % use 8-bit T1 fonts
\usepackage[hidelinks]{hyperref}       % hyperlinks
\usepackage{url}            % simple URL typesetting
\usepackage{booktabs}       % professional-quality tables
\usepackage{amsfonts}       % blackboard math symbols
\usepackage{nicefrac}       % compact symbols for 1/2, etc.
\usepackage{microtype}      % microtypography
\usepackage{graphicx}

\usepackage{EvolvingAIcommenting}
\newif\ifcomments

\commentstrue

\ifcomments
\newcommand{\comments}[1]{#1}
\else
\newcommand{\comments}[1]{}
\fi
\title{On the Relationship Between the OpenAI Evolution Strategy and Stochastic Gradient Descent}

\author{
  Xingwen Zhang, Jeff Clune, Kenneth O. Stanley \\
  Uber AI Labs\\
  San Francisco, CA 94103 \\
  \texttt{\{xingwen,jeffclune,kstanley\}@uber.com} \\
}

\begin{document}

\maketitle
\begin{abstract}
%Deep learning methods based on stochastic gradient descent (SGD) have provided a major boost to performance in reinforcement learning (RL).  
Because stochastic gradient descent (SGD) has shown promise optimizing neural networks with millions of parameters and few if any alternatives are known to exist, it has moved to the heart of leading approaches to reinforcement learning (RL).  For that reason, the recent result from OpenAI showing that a particular kind of evolution strategy (ES) can rival the performance of SGD-based deep RL methods with large neural networks provoked surprise.  This result is difficult to interpret in part because of the lingering ambiguity on how ES actually relates to SGD.  The aim of this paper is to significantly reduce this ambiguity through a series of MNIST-based experiments designed to uncover their relationship.  As a simple supervised problem without domain noise (unlike in most RL), MNIST makes it possible (1) to measure the correlation between gradients computed by ES and SGD and (2) then to develop an SGD-based proxy that accurately predicts the performance of different ES population sizes.  These innovations give a new level of insight into the real capabilities of ES, and lead also to some unconventional means for applying ES to supervised problems that shed further light on its differences from SGD.  Incorporating these lessons, the paper concludes by demonstrating that ES can achieve 99\% accuracy on MNIST, a number higher than any previously published result for any evolutionary method. While not by any means suggesting that ES should substitute for SGD in supervised learning, the suite of experiments herein enables more informed decisions on the application of ES within RL and other paradigms. 
\end{abstract}

\section{Introduction}

%KENLAST: get CPU/GPU usage and time in days

The recent revelation from OpenAI \citep{es} that an evolutionary algorithm can rival modern gradient-based reinforcement learning (RL) approaches in the Mujoco and Atari learning environments is a source of some surprise.  Perhaps the most surprising part is simply that a search algorithm that does not compute exact gradients can optimize deep neural networks (DNNs) with over a million parameters.  Until recently, evolution was thought to falter in high-dimensional spaces (e.g\ above 10,000 parameters), but in \citet{es} it seems unimpeded by hundreds of thousands and even millions.  This paper is an attempt to make sense of this recent result by showing more clearly how an OpenAI-style evolutionary algorithm is related to stochastic gradient descent (SGD), thereby providing greater clarity to researchers uncertain about the implications of choosing one or the other paradigm.  

Evolving neural networks, more commonly called \emph{neuroevolution}, has a long history preceding the recent \citet{es} result \citep{floreano:ei08,yao:ieee99}.  In fact, it is important to note that over the history of RL, there are periods during which neuroevolution was the leading method on popular benchmarks.  For example, the NeuroEvolution of Augmenting Topologies (NEAT) algorithm \citep{stanley:ec02} was a leader in a number of pole balancing variants in the early aughts even compared to leading conventional RL approaches of the time.  Other neuroevolution methods such as CoSyne \citep{gomez:ecml06} and CMA-ES \citep{igel:cec03} also returned top-tier results in subsequent years.  Hypercube-based NEAT (HyperNEAT), a successor algorithm to NEAT based on indirect encoding \citep{stanley:alife09,gauci:nc10}, later also exceeded the performance conventional RL algorithms in another popular benchmark of its time, Keepaway Soccer \citep{verbancsics:jmlr10}.  There are even recent hints that a special kind of neuroevolution algorithm  called a \emph{limited perturbation evolutionary algorithm} (LEEA) can sometimes match and exceed the performance of SGD in low-dimensional networks \citep{morse:gecco16}.

However, the ability of neuroevolution to compete at the cutting edge of RL has become less clear as deep learning has eclipsed the capabilities of numerous rival approaches across many domains, to a large extent by finally showing how very large and deep networks can be trained reliably.  The old neuroevolution results were often on networks orders of magnitude smaller than those trained in deep learning today.  HyperNEAT is an exception because its indirect encoding means it can evolve much large networks (e.g. over 100,000 weights), but its networks, such as in Keepaway, were still relatively shallow.  So while it is not documented explicitly in the literature, a belief seems to have set in that while evolution has some interesting capabilities in low-dimensional neural optimization for RL, deep learning has largely moved us into a new era of larger and deeper networks.

The result from \citet{es} has potentially undermined this assumption by showing that evolution and large networks are actually compatible.  Because of the significant implications of this result, it is important to clear up some potential confusion about the approach:  First, the name chosen to refer to the algorithm in the OpenAI work -- an \emph{evolution strategy} (ES) -- historically refers to a wide class of algorithms going back to the 1970s that are loosely inspired by evolution and that often involve generating more offspring than parents (with as few as a single parent in one generation) \citep{beyer:nc02}.  That is, a small set of parents generates a larger set of offspring, from which the next (small) set of parents are then assembled.  The OpenAI-ES is related to a single-parent ES, but it is closest to the more recent \emph{natural evolution strategy} (NES) variant \citep{wierstra:jmlr14}.  An important idea in the NES is to extract useful information from \emph{all} the offspring, including those performing poorly.  
The NES in effect keeps track of a current \emph{distribution} of solutions and treats offspring as samples that are aggregated to adjust the overall offspring distribution (both in mean and standard deviation) in further steps.  The OpenAI-ES then is in effect a simplified NES that only adjusts the mean, keeping the shape of the distribution around the mean consistent.  A consequent appeal of this approach is its simplicity.  

For simplicity and consistency with the literature, we refer to the OpenAI-ES from this point forward as ES, with the understanding that in this case the subject is the specific ES variant investigated by \citet{es}. 

%TODO: fix quotation marks
Because ES (like NES) aggregates an entire population of offspring to decide the next step for the algorithm, some view it as closer to a finite-difference approximator than an evolutionary algorithm in its typical realization \citep{ebrahimi:arxiv17}.  
%Perhaps for some this interpretation is even more comforting given that it might help to maintain a consistent assumption that evolution cannot effective optimize millions of parameters consistent.   
However, while indeed evolution in nature does not aggregate (i.e. ``mate'' all at once) hundreds or thousands of individuals to generate a new point in the search, ES is still not a simple finite difference-based gradient approximator (as it is e.g.\ described by \citet{ebrahimi:arxiv17}):  As analyzed by \citet{wierstra:jmlr14}, it optimizes not just a point but a \emph{whole population}, which means that it actually does maintain some properties conceptually distinct from gradient descent.  We demonstrate the implications of optimizing a population versus a point in a companion paper \citep{lehman:arxiv17fd}.
Another benefit of the name ES is to connect it to the long historical precedent to this algorithm that stems from the evolutionary computation literature.  Whether or not it is truly evolution in the spirit of nature is probably not as important for machine learning as what it actually does and how it does it, which is the focus of this work.

To that end, we answer a series of questions in this paper aimed at understanding how and when ES can rival SGD-based RL algorithms.  Across RL methods, the gradient provides an imperfect signal for obtaining optimal performance.  Environmental noise is a factor in degrading the utility of the gradient, in addition to the potential for converging to local optima from greedily maximizing local reward.  For these reasons, learning in RL generally requires multiple trials or \emph{rollouts}.  Ideally, these rollouts could be completed in parallel, but for SGD-based approaches, that requires sending gradient vectors of potentially millions of dimensions between processors.  In contrast, in ES, because rollouts correspond to perturbations generated from a random seed, only the random seeds and scalar reward values need to be shared over the computer network, making parallelism more practical.  In this way, as argued by \citet{es}, ES offers a potential practical advantage in parallelization compared to other RL techniques.  ES also offers other potential advantages, such as its tendency to seek out more robust areas of the search space \citep{lehman:arxiv17fd}, but here we focus on its gradient-following capability and the impact of that capability on performance.  A better understanding of this relationship can help to motivate why ES is a justifiable option for high-dimensional problems in RL.  

To reach this understanding, the core approach is to study the behavior of ES empirically in a high-dimensional context (over three-million weights) removed from the noise, lack of ground truth, and other confounds present in the world of RL.  
That is, while we are motivated by applications in RL, the added layers of complexity in RL as opposed to supervised learning can
make it difficult to tease out the underlying gradient-following capabilities of ES, which this paper makes more clear.
While in the past the MNIST benchmark \citep{lecun:mnist98} has served as a gauge for progress in the field, here it serves the different purpose of supporting several variant experiments in which it is often possible to measure the correlation between the gradient calculated by SGD and approximated by ES, thereby providing a clearer picture of what ES is actually doing compared to SGD.  The relationship of ES to SGD is further clarified by the introduction of a novel proxy for ES based on SGD with varying levels of noise that models the performance of ES with different population sizes.  Through this proxy, we can predict the necessary level of computation in ES not only for reaching various levels of performance, but also for effectively matching the performance of SGD.

With these baselines established, it becomes easier to understand why ES might be competitive in domains like RL where indirect access to the performance gradient degrades the ideal conditions often available to SGD in supervised problems. In particular, an important lesson is that even with what seems a low correlation between an approximated gradient and the optimal gradient computed for SGD, it is still possible to achieve surprisingly high-level performance, in part explaining why SGD may lose its competitive edge when its own ideal gradient is degraded in RL.  

Finally, we introduce some novel tricks for increasing the performance of ES (both in terms of speed and gradient approximation quality) in practice that yield further insight into its function.  To demonstrate the consequence of these insights in practice, a final experiment obtains a respectable 99\% performance on MNIST through ES alone, a level not previously reported through evolutionary means.  This performance level hints at a likely increase in the applicability of ES to deep RL problems as available processing power increases.
Ultimately, the hope of this paper is to further inform decision-making in the future as the toolbox of available algorithms for training large networks expands.  
%KEN(idea): Could we actually create a very simplified problem that is like an RL problem where we could see SGD degrade to the point of parity with ES?  What would that look like?

%TODO: note the proxy idea also in the intro

%-is it an EA?

%-what should it be called?

%-relating MNIST to RL

\section{Background}

Starting from an initial set of parameters, denoted by $\theta_t$ at iteration $t$, the OpenAI-ES generates $N$ \emph{pseudo-offspring} by perturbing $\theta_t$ with a random noise vector $v_{t,i}$ where $1 \leq i \leq N$. Each component of the noise vector $v_{t,i}$ is independently sampled from a normal distribution with mean 0 and standard deviation of $\sigma$. The term pseudo-offspring helps to distinguish the process in ES whereby all the population is aggregated back into a single point on each iteration from the more conventional case in evolutionary algorithms whereby offspring reproduce to form a next-generation population \citep{dejong:book02}. Each pseudo-offspring, i.e., the parameters $\theta_t + v_{t,i}$, is evaluated to compute its reward $r_{t, i}$, which could be a raw reward or a center-ranked reward \citep{es}. After collecting the rewards for all the pseudo-offspring, the gradient is estimated as
\begin{equation}
g_{t}^{\textrm{\small ES}} = \frac{1}{N\sigma^2}\sum_{i=1}^{N} v_{t,i} r_{t,i}, \label{gES}
\end{equation} and the parameters are then updated using SGD as
\begin{equation}
\theta_{t+1} = \theta_t + \alpha g_{t}^{\textrm{\small ES}},
\end{equation}
where $\alpha$ is the learning rate. Other optimizers, such as ADAM \citep{es}, can also be applied, again using $g_{t}^{\textrm{\small ES}}$ as the gradient for the $t$-th iteration.
%DONE KENTODO: Make sure the above equations are consistent with Salimans.  Jeff says it seems some terms may be missing (maybe we folded them into the learning rate?)
%DONE KENTODO: What does "up to a scaling factor" mean above?

For MNIST, we define the reward of a pseudo-offspring as the negative of the cross-entropy loss when it is evaluated on a set of images, i.e., a mini-batch.

OpenAI-ES is very scalable because the computation of each $r_{t,i}$ can be distributed among a large number of workers. Furthermore, the workers do not send the noise perturbation $v_{t,i}$ to the master, which is large for large DNNs; instead, the master and the workers share a common noise table, and thus it suffices for the worker to send the index of $v_{t,i}$ in the noise table to the master.

\section{Approach: Computing Gradient Correlations}

%HERE

Interestingly, by investigating ES in the context of supervised learning, it becomes possible to compare its gradient estimates to the exact gradients computed by SGD. In particular, for any given mini-batch of training examples provided to SGD during learning, the outputs of the DNN can be compared to the correct targets (i.e.\ optimal gradients from backpropagation \citep{rumelhart:errorprop}) to derive changes in weights to improve DNN performance with respect to that mini-batch in the future.  In practice, this gradient is computed analytically or numerically using tools such as Tensorflow. Let us denote the true gradients by $g_{t}^{\textrm{\small True}}$.
%KENTODO: decide on use of \textrm{hello} in equations - YES )(Xingwen)

Given a particular mini-batch in a supervised problem domain, instead of computing the exact gradient, ES computes an approximation from all the sample points (called pseudo-offspring) generated from parent $\theta_t$.  Recall that pseudo-offspring are generated by stochastically perturbing $\theta_t$ from a normal distribution, and the estimated gradients $g_{t}^{\textrm{\small ES}}$ are given in the previous section.

It is important to note that in the Mujoco and Atari results from \citet{es}, the fitness (i.e.\ performance) values of pseudo-offspring are \emph{rescaled} through a centered-rank procedure. This procedure ranks all raw rewards and then scales the ranks to a given range (for example, [-0.5, 0.5]). Unlike in the RL experiments from \citet{es}, MNIST has no noise from rollouts.  Therefore, the unbiased estimate of  the gradient provided by raw reward in equation \ref{gES} has fewer sources of potential variance.  For this reason, we simply use the raw reward to estimate the gradients.
%DONE KENTODO: Need to explain better why can use raw reward when noise is not present %HERE

Given a mini-batch, the optimizer always uses $g_{t}^{\textrm{\small ES}}$ to update the parameters $\theta_t$. Meanwhile, for the same mini-batch we can also compute $g_{t}^{\textrm{\small True}}$, and the correlation between $g_{t}^{\textrm{\small ES}}$ and $g_{t}^{\textrm{\small True}}$ (denoted by $\rho_t$) as 
\begin{eqnarray}
\rho_t &=& \textrm{correlation}(g_{t}^{\textrm{\small ES}}, g_{t}^{\textrm{\small True}}) \nonumber \\
&=& \frac{\sum_{i=1}^{P}{(g_{t, i}^{\textrm{\small ES}} - \bar{g}_t^{\textrm{\small ES}})(g_{t, i}^{\textrm{\small True}} - \bar{g}_t^{\textrm{\small True}})}}{\sqrt{\sum_{i=1}^{P}{(g_{t, i}^{\textrm{\small ES}} - \bar{g}_t^{\textrm{\small ES}})^2}} \sqrt{\sum_{i=1}^{P}{(g_{t, i}^{\textrm{\small True}}  - \bar{g}_t^{\textrm{\small True}})^2}}},
\end{eqnarray}
where $P$ is the number of parameters and $\bar{g}_t^{\textrm{\small ES}}$ and $\bar{g}_t^{\textrm{\small True}}$ are the means of $g_t^{\textrm{\small ES}}$ and $g_t^{\textrm{\small True}}$, respectively. To focus the measurements on the performance of ES, unless explicitly stated otherwise, $g_{t}^{\textrm{\small True}}$ is not used by the optimizer.  That is, SGD gradients are always computed from the current ES $\theta_t$ so they can be compared with the gradient estimated by ES from that same $\theta_t$.

As a point of reference for understanding the implications of different correlation levels, it is interesting to analyze how SGD correlates to itself given different random mini-batches. For this purpose, in each iteration, we fetch two random mini-batches and calculate $g_{t}^{\textrm{\small True}}$ for each mini-batch, and a correlation between them can then be computed.  This correlation in MNIST is on average 39.04\%.

Thanks to this ability to compute the correlation between the exact and approximate gradient for a mini-batch, it becomes possible to answer fundamental questions about the relationship between the true gradient and the approximation from ES under different conditions, as attempted next.

\section{The OpenAI-ES on MNIST}

%TODO: Do we state the perturbation magnitude for the ES anywhere (sigma)?

With the ability to calculate the correlation between gradients computed by ES and SGD, MNIST becomes a convenient testbed for understanding the raw behavior of ES outside the confounds of RL.  These confounds include deception, wherein the gradient computed by greedily pursuing reward can be misleading, and also the environmental noise that can occur in RL.  Reflecting our interest in the performance of evolution in high-dimensional spaces, the networks optimized in these experiments have 3,274,634 parameters.  More specifically, the network (as taken from the Tensorflow tutorial at http://www.tensorflow.org/get\_started/mnist/pros) has four layers, including two convolutional layers with 32 and 64 kernels per conv layer respectively (both with stride of 5), followed by ReLU activation functions and 2-by-2 max pooling. The last two layers are fully connected with 1,024 and 10 units respectively. The mini-batch size is 50 and the optimizer is ADAM with a fixed learning rate of 0.001. Lastly, in all experiments $\sigma$ is fixed at 0.002, which was found most effective through preliminary experiments and whose low value makes sense because of the lack of environmental noise in MNIST.
%KENTODO: What is the stride of convolutions here?

The correlation between ES and SGD depends upon the number of pseudo-offspring produced for each iteration of ES.  In the limit as the number of pseudo-offspring increases, the gradient approximation of ES converges to the analytic gradient when the noise perturbations are small.  However, the key question is how this relationship works in practice.  For example, in the code released with \citet{es} the number of pseudo-offspring is 10,000.  This number is interesting because as CPU costs decrease, the ability to evaluate 10,000 candidates quickly is realistic even today (which reflects one of the advantages touted for ES in RL).  Large commercial organizations today often possess well beyond 10,000 CPUs, making complete parallelism on such a population size a practical option if advantageous.  

Note that in this investigation we routinely measure the accuracy of the current $\theta_t$ across the entire MNIST test set of 10,000 examples.  Because these measurements are taken at every iteration (i.e.\ not just the end of training) we call this measurement the \emph{validation accuracy} throughout the text, but it is important to emphasize for clarity that it includes the entire test set.  

\begin{figure}[h]
  \centering
  \includegraphics[width=\linewidth]{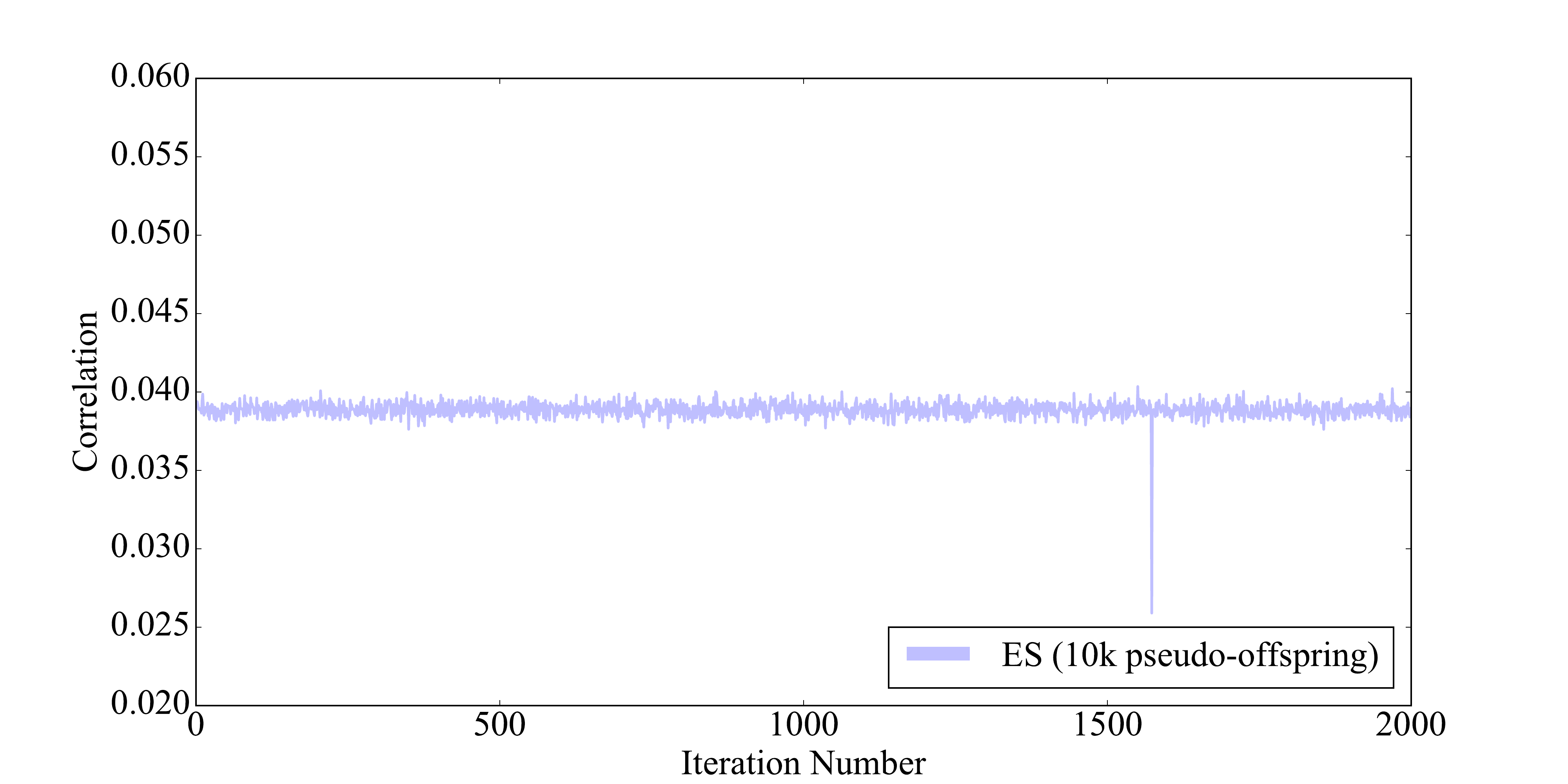}
  \caption{\textbf{Correlation of the gradients estimated by ES and the analytic gradients for the same sequence of mini-batches}.  The correlation between ES and gradients used by SGD is remarkably stable.}
  \label{correlation}
\end{figure}

It turns out that the average correlation over an entire 2,000-iteration run of ES (while simultaneously computing gradients for SGD) with 10,000 pseudo-offspring is 3.9\%.  This correlation is remarkably stable across iterations (figure \ref{correlation}).  The natural next question is whether a 3.9\% correlation to the optimal gradient is sufficient to learn MNIST.  After 2,000 iterations ES with this correlation level achieves an average validation accuracy of 96.99\% across 5 runs, compared to 98.69\% achieved by SGD for this architecture over the same number of iterations, where an iteration corresponds to one update to $\theta_t$ by the optimizer.  
%DONE KENTODO: Should we define "iterations" for SGD?
This performance discrepancy gives a sense of the cost of imperfect correlation to the optimal gradient; however, while certainly a significant cost is exacted, it is perhaps surprising that 3.9\% correlation still comes within 1.70\% of the validation accuracy by SGD.

%KENTODO: in future version of paper we could also include a plot showing learning curve of SGD next to ES

%KENNOTE: note Zoubin had a suggestion here regarding: "looking at the distribution of the dot product between the two gradients, how often is it positive vs. negative"
%They are all positive.

%KENLAST: fix use of words validation versus test??
While providing a hint of the approximating power of ES, the unsatisfying aspect of this result is that it covers only a specific case.  More informative would be a comprehensive study that shows both what level of correlation and how many pseudo-offspring are needed to match the performance of SGD.  Yet performing a comprehensive survey of pseudo-offspring population sizes is currently prohibitively computationally expensive, especially as population sizes begin to exceed 100,000.  As an alternative to running such a study, we propose the intriguing alternative that it is possible to construct an SGD-based \emph{proxy} for ES that runs much faster.  This proxy also yields further insight into the relationship between SGD and ES; in particular, the hypothesis is that for ES with any population size $N$ of pseudo-offspring and low $\sigma$, there is an equivalently-performing version of SGD with some level of independent noise added to it that produces the same correlation of the gradients. Denoting the perturbed gradients by $g_{t}^{\textrm{\small Proxy}} = g_{t}^{\textrm{\small True}} + m g_{t}^{\textrm{\small Noise}}$, where $g_{t}^{\textrm{\small Noise}}$ is the independent noise vector, we set the multiplier $m$ such that the distribution of $\textrm{correlation}(g_{t}^{\textrm{\small Proxy}}, g_{t}^{\textrm{\small True}})$ closely matches the distribution of $\textrm{correlation}(g_{t}^{\textrm{\small ES}}, g_{t}^{\textrm{\small True}})$. More specifically, to achieve a desired level of correlation $\rho$, we have the corresponding multiplier $m=\sqrt{\frac{\textrm{variance}(g_{t}^{\textrm{\small True}})(1 - \rho^2)}{\textrm{variance}(g_{t}^{\textrm{\small Noise}})\rho^2}}$, such that $\textrm{correlation}(g_{t}^{\textrm{\small Proxy}}, g_{t}^{\textrm{\small True}})$ is then $\rho$.  The multiplier $m$ decreases monotonically as $\rho$ increases; intuitively, if you want higher $\rho$, then you need a smaller multiplier $m$. Note that augmenting SGD with noise in this way somewhat resembles stochastic gradient Langevin dynamics \citep{welling:icml11}, but the motivation for this ES proxy differs significantly from that method.
%DONE KENTODO: Need to address Jeff's comments in the math at the end of the above paragraph, and also explain the equation for m intuitively.

\begin{figure}[h]
  \centering
  \includegraphics[width=\linewidth]{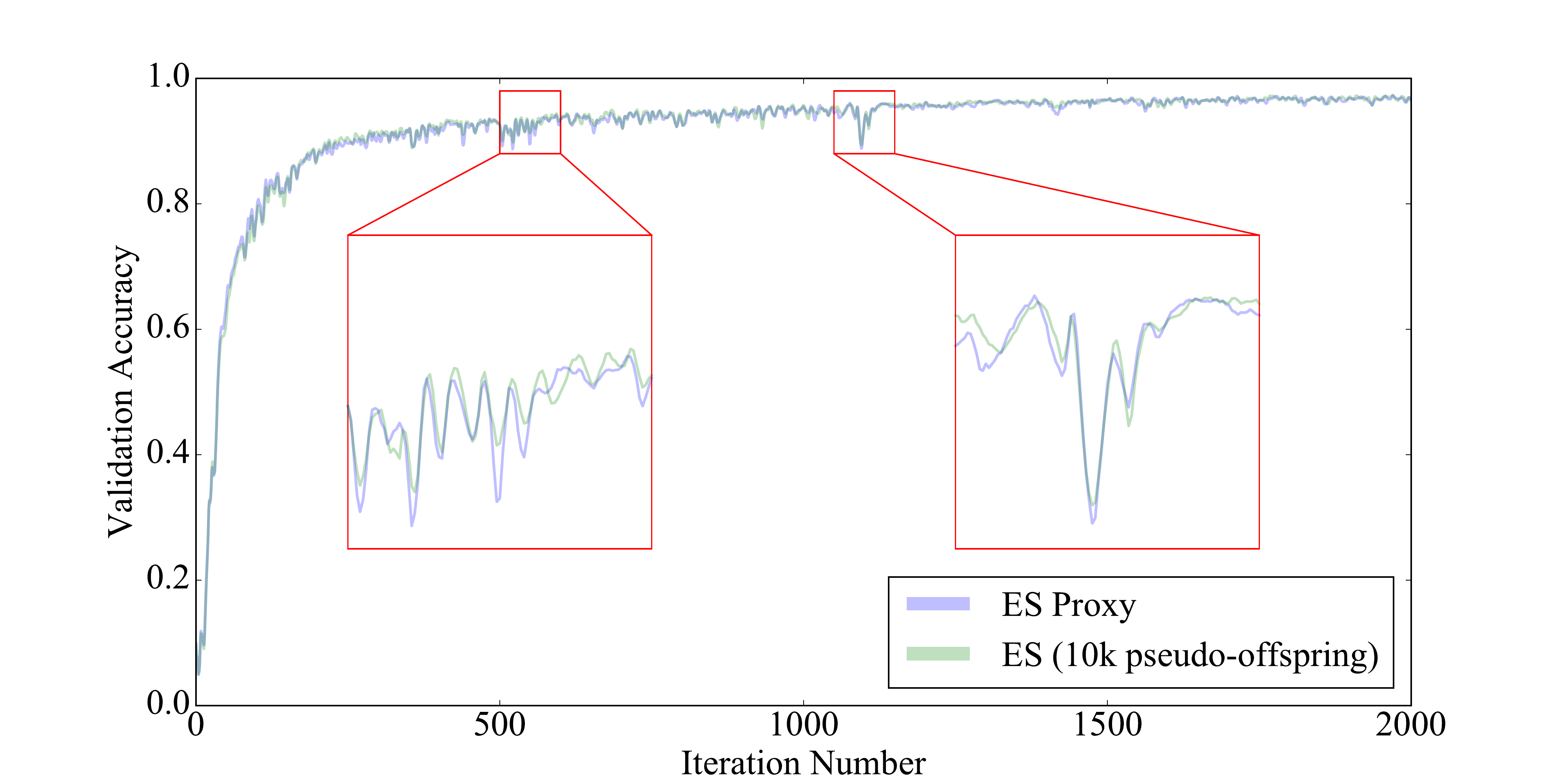}
  \caption{\textbf{Validation accuracy of ES and the ES proxy for the same sequence of mini-batches.} Notice how the fluctuations in performance by ES and its proxy are nearly identical throughout the run despite the randomness in the SGD proxy and the ES pseudo-offspring.  Insets show local areas of the curves close up.  The implication is that mini-batches are the primary driver of noise in the search, impacting ES and SGD in the same way.  Recall that validation accuracy over iterations in this figure and throughout this paper is reported on the entire MNIST test set.} 
  \label{proxy:accuracy}
\end{figure}

%DONE KENTODO: say something about why we use small sigma in this study? (as compared to Salimans?)
Although it may not be obvious that this hypothesis on the equivalence between SGD with noise and ES holds, it turns out that the correlation between ES and an SGD-based proxy is striking (at least with the small $\sigma$ used in this study, which makes ES similar to a finite-difference approximation).  For example, for the same sequence of mini-batches, the learning curve for the ES proxy (SGD with appropriately scaled noise) is almost \emph{identical} to ES with $n=10,000$ pseudo-offspring (figure \ref{proxy:accuracy}).  Notice how even fluctuations in validation accuracies between ES and its proxy are almost identical, suggesting that the gross fluctuations in the learning curve are from the mini-batches and not from the noise in the generation of pseudo-offspring.  In this way, the ES proxy is almost a perfect model for ES.  
Not only does this insight now allow us to investigate the implications of different correlation levels on ultimate performance, but it also provides an important further insight into the nature of the gradient approximation by ES with low $\sigma$: in effect it confirms that it is nearly identical to running SGD with a certain level of noise, even at a fine-grained level (figure \ref{proxy:accuracy}). 
%NOTE: can't include below footnote unless/until other paper is ready
%\footnote{Note that a companion paper \cite{TODO-jay} also shows an important \emph{difference} between ES and SGD, which is that in stochastic domains ES also tends to find solutions that are more robust to parameter perturbations.}

With this SGD-based ES proxy, it is now possible to predict the ultimate expected performance level for a given level of gradient correlation by ES.  That is, given a sufficient number of mini-batches (e.g.\ 40,000), how well would versions of ES with different levels of correlation perform?  Table \ref{proxy_table} shows this relationship and compares it to the SGD control: a correlation of 20\% is sufficient to achieve the validation accuracy of 99\% (averaged over 20 runs) after 40,000 iterations.  Note that this number of iterations is greater than in the previous experiment without the proxy -- because the proxy can be tested so quickly, it can be run for more iterations over multiple runs to expose its long-term average convergence level.
A natural question then arises: how many pseudo-offspring are needed to achieve a correlation of 20\%? To answer this interesting question, we let the workers continue to evaluate pseudo-offspring for a given set of parameters, while the master re-estimates the gradients for all  pseudo-offspring whenever it receives 1,000 additional pseudo-offspring evaluations, stopping when the estimated gradient is 20\% correlated to the true gradient. It turns out that the level of 20\% correlation emerges at about 274,000 pseudo-offspring, thus giving a sense of where SGD-performance equivalence occurs in MNIST. 

\begin{table}[t]
  \centering
  \begin{tabular}{lrr}
    \toprule
    Name             &  mean(Validation Accuracy)  & std(Validation Accuracy)  \\
    \midrule
    SGD              &  99.11\%                      & 0.10\%  \\
    ES Proxy (4\% correlation)   &  98.39\%                      & 0.13\%  \\
    ES Proxy (20\% correlation)  &  98.99\%                      & 0.11\%  \\
    \bottomrule
  \end{tabular}
  \vspace{0.1in}
    \caption{\textbf{Predicted ES performance after 40,000 iterations based on proxies with different correlation levels.} The results are averaged over 20 runs. SGD-equivalence appears to occur at about 20\% correlation to the analytic gradient.}
  \label{proxy_table}
\end{table}

The results so far establish that ES is similar to a noisy version of SGD, and that noise in ES can be reduced with more pseudo-offspring.  They also show that even with what seem like low levels of correlation (e.g.\ 3.9\%), optimization can still reach surprisingly high levels of performance.  The totality of results is interesting because while the computational effort needed to reach parity may not be worthwhile in supervised domains, in problems like RL where gradient computation is a less reliable compass to higher performance, the fact that low levels of correlation nevertheless reach surprisingly high performance in supervised tasks begins to signal an opportunity.
%the hardware requirements to optimize even at nearly SGD-equivalent levels are %obtainable today and will only become more so over time. 

ES benefits from massive CPU parallelism, requiring a different kind of infrastructure investment, whose costs and benefits will play out with the changing prices and
shifting demands of hardware for different tasks.
Furthermore, as argued by \citet{es}, the ability to pass random seeds corresponding to pseudo-offspring perturbations as opposed to entire high-dimensional gradients means that ES in practice can benefit more from parallelization in e.g.\ RL tasks, because large numbers of parallel gradient adjustments of millions of parameters can be passed inexpensively (by passing the seed that produces the high-dimensional perturbation rather than the entire parameter vector itself) to a master node that then aggregates all of them. 
In conventional RL, on the other hand, gradient vectors would need to be passed to the master in their entirety, because there is no seed that generated them. 
This story offers a more informed perspective on ES, but (as discussed in detail in the next section) it is also important to recognize that it does not address one expensive bottleneck for ES that does get worse with increasing numbers of pseudo-offspring: In particular, there is the 
prohibitive expense at the master itself of combining thousands of million-dimensional vectors, which detracts to some extent from the overall positive implications of parallelism.

\subsection{Limited Perturbation ES}

For a network having as many as 3,274,634 parameters, the master can be a bottleneck when all parameters are perturbed for each pseudo-offspring, even despite the advantages of ES for worker parallelism (an issue also noted by \citet{es}). The bottleneck results from the master's need to combine thousands of perturbation vectors, each of potentially millions of dimensions, to compute its final gradient step.
In this way, the overall computation time to estimate $g_t^{\textrm{\small ES}}$ as in equation \ref{gES} increases linearly within the master with the number of parameters in the network.  
For example, computing the final aggregate gradient from 10,000 pseudo-offspring with 3,274,634 parameters each takes roughly 90 seconds in ES on an Intel(R) Xeon(R) CPU of 2.20GHz.
As mentioned in \citep{es}, the master could aggregate perturbations much faster if each pseudo-offspring only perturbed a subset of the current $\theta_t$.  However, no study to date has examined how such limited perturbation impacts the gradient estimating abilities or computational speed of ES. For this purpose, we ran a version of ES where the parameters are partitioned into a number of buckets, each of which contains a subset of up to 5,000 parameters. Each pseudo-offspring then perturbs only parameters in a randomly chosen bucket. The reward of each pseudo-offspring is computed and sent to the master together with the corresponding noise index as before and the index of the chosen parameter bucket. For any pseudo-offspring, the master then only aggregates over the parameters that are actually perturbed, ignoring those unperturbed ones, and the computation effort in the master is thus greatly reduced.

\begin{figure}[h]
  \centering
  \includegraphics[width=\linewidth]{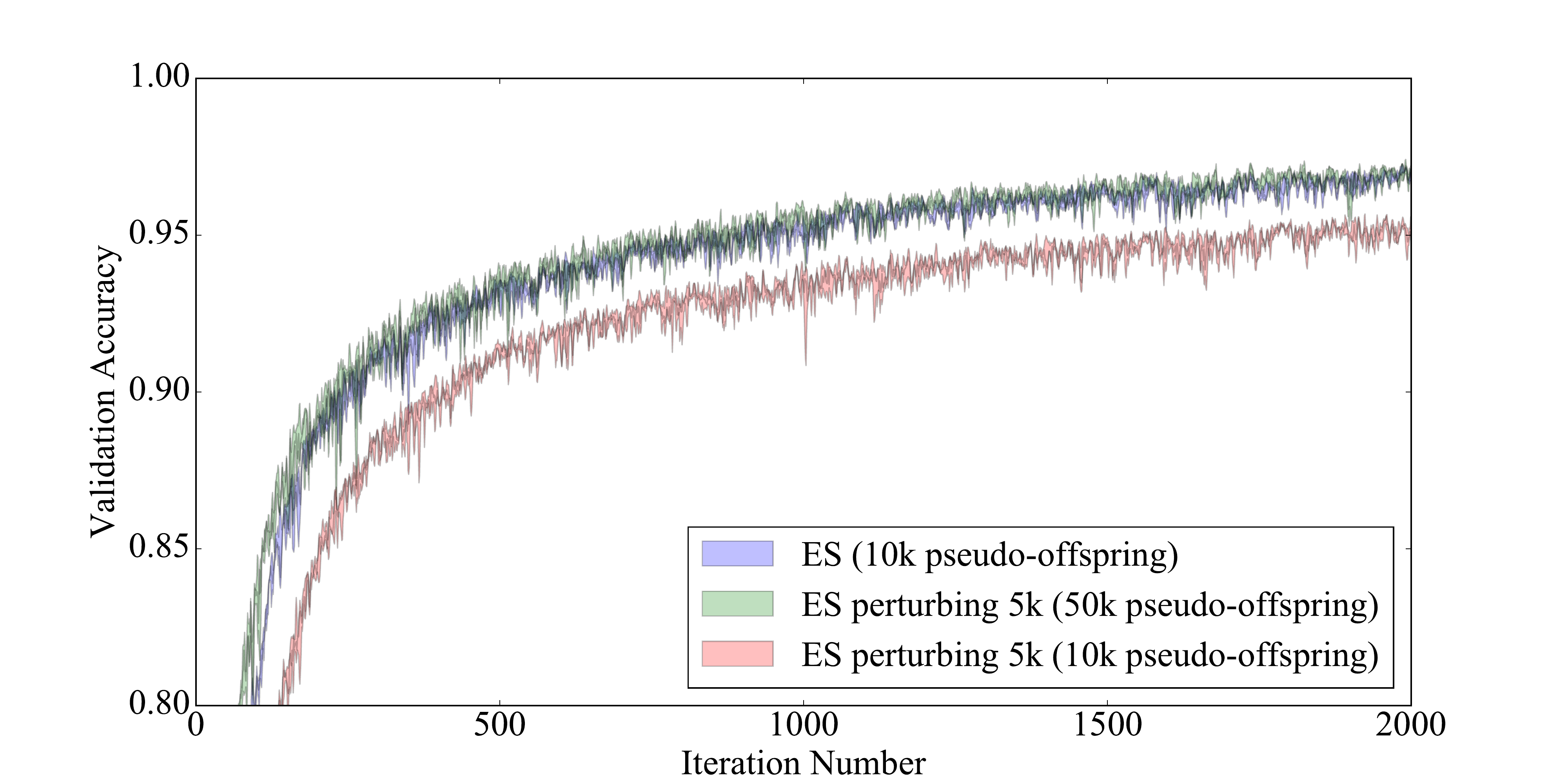}
  \caption{\textbf{Comparing the performance of ES with and without limited perturbation.} For each method, the median validation accuracy across 5 runs on MNIST is plotted and the area between the 25th and 75th percentiles is filled. ES with 50,000 pseudo-offspring each perturbing 5,000 parameters was able to achieve equivalent performance to ES without limited perturbation (i.e.\ their difference is statistically insignificant; $p=0.75$ from the Wilcoxon test). Because there is no significant difference between these two curves, they are seen above overlapping in the plot.
 %\jc{Means and standard deviations assume a normal distribution, and you have to justify that assumption or conduct tests to prove the data are normal in order to use them. Medians and CIs are safer/easier for that reason. Also, I am nearly certain that it is improper to claim something is significant just by eyeballing it and saying the SDs do not overlap. Doing that is an informal heuristic, but is not acceptable in scientific publications. I recommend a Mann Whitney of the final performances and reporting a p value. If you want p values over time, see recent Huizinga papers out of my lab. The mean/std mistake is commonly made, and, while incorrect, is thus not as big of a deal, but the eyeballing statistical significance is something we definitely need to fix}\todo{we talk about one comparison, but not the others. it's not strictly required, but might be nice to say whether ES 5k is significantly lower than the other two} 
 }
  \label{es:limited-perturbation}
\end{figure}

As shown in figure \ref{es:limited-perturbation}, ES with 50,000 pseudo-offspring each perturbing 5,000 parameters is able to achieve equivalent performance as regular ES with 10,000 pseudo-offspring perturbing all parameters. 
In contrast to the 90 seconds taken by the master to complete its gradient estimate based on 10,000 pseudo-offspring, the limited perturbation ES (perturbing 5,000 parameters per pseudo-offspring)  takes instead just 1.4 seconds when aggregating 50,000 pseudo-offspring, which is a significant speed-up of \emph{64 times}, although not achieving the theoretical speed-up ratio of $\frac{3274634 \times 10000}{5000 \times 50000} \simeq 131$ due to overhead.  Thus, although requiring five times more pseudo-offspring, limited perturbation ES is able to obtain the same final performance, and can run much faster than a regular ES with fewer pseudo-offspring  if a large number of workers are available.  For example, in our experimental setup, 100 workers in regular ES with 10,000 pseudo-offspring took approximately twice as long to complete a full run than 500 workers with limited perturbations and 50,000 offspring.
%KENLAST: how long is twice as long?

\subsection{No-mini-batch ES}

A further interesting insight is that there is in principle no need to divide the training set into mini-batches for ES the way we do conventionally for SGD.  
%Mini-batches are important for SGD in part because they are a source of stochasticity, helping to avoid some local minima and of course greatly reducing the memory footprint.\jc{There are other reasons, and somehow this makes us come off as naive and not understanding SGD. For example, I think a major benefit is faster learning owing to not needing accurate gradients early in training, and thus taking many noisy steps is better than a few accurate ones because you are so far from the goal.}
%\todo{could this approach work with SGD too? you can still do it with small batch size, which gets at the memory issue. You can also get stochasticity because each gradient sample would be over a different mini-batch. Does something about GPU computing make this inefficient? More importantly, has someone already done this for SGD? We may want to ask around our lab. KEN: Not clear what "this approach" would mean for SGD.}  
Because in ES there is already stochasticity from the generation of the pseudo-offspring, there is not necessarily a need for additional stochasticity through mini-batches.  Nevertheless, because it would be expensive, it is still desirable to avoid evaluating each pseudo-offspring on the entire training set.  The overall implication is that instead of dividing the training set into a separate mini-batch for each iteration, \emph{each pseudo-offspring in every iteration} can be evaluated against a unique random subset of the whole training set.  This \emph{no-mini-batch ES} might actually be expected to outperform conventional mini-batches because the step taken at each iteration of the ES is calculated from a more comprehensive sampling of the whole training set (when taking into account the aggregation of all the pseudo-offspring).  This insight is relevant to domains without direct supervision like RL because they do not proceed by mini-batches, so a mini-batch-based measure of the capabilities of ES as in the MNIST experiments reported on so far might be viewed as pessimistic.  In this way, the no-mini-batch approach could be more reflective of the true potential of ES for gradient following in RL. 
%\jc{Also note that sometimes we do the ES and sometimes just ES...given limited time this is a minor issue, but perhaps one to fix down the road (in Catalyst and Horizons papers, because we do it too!)}.
%TODO: cite Morse 2016 paper here about limited evaluation

\begin{figure}[h]
  \centering
  \includegraphics[width=\linewidth]{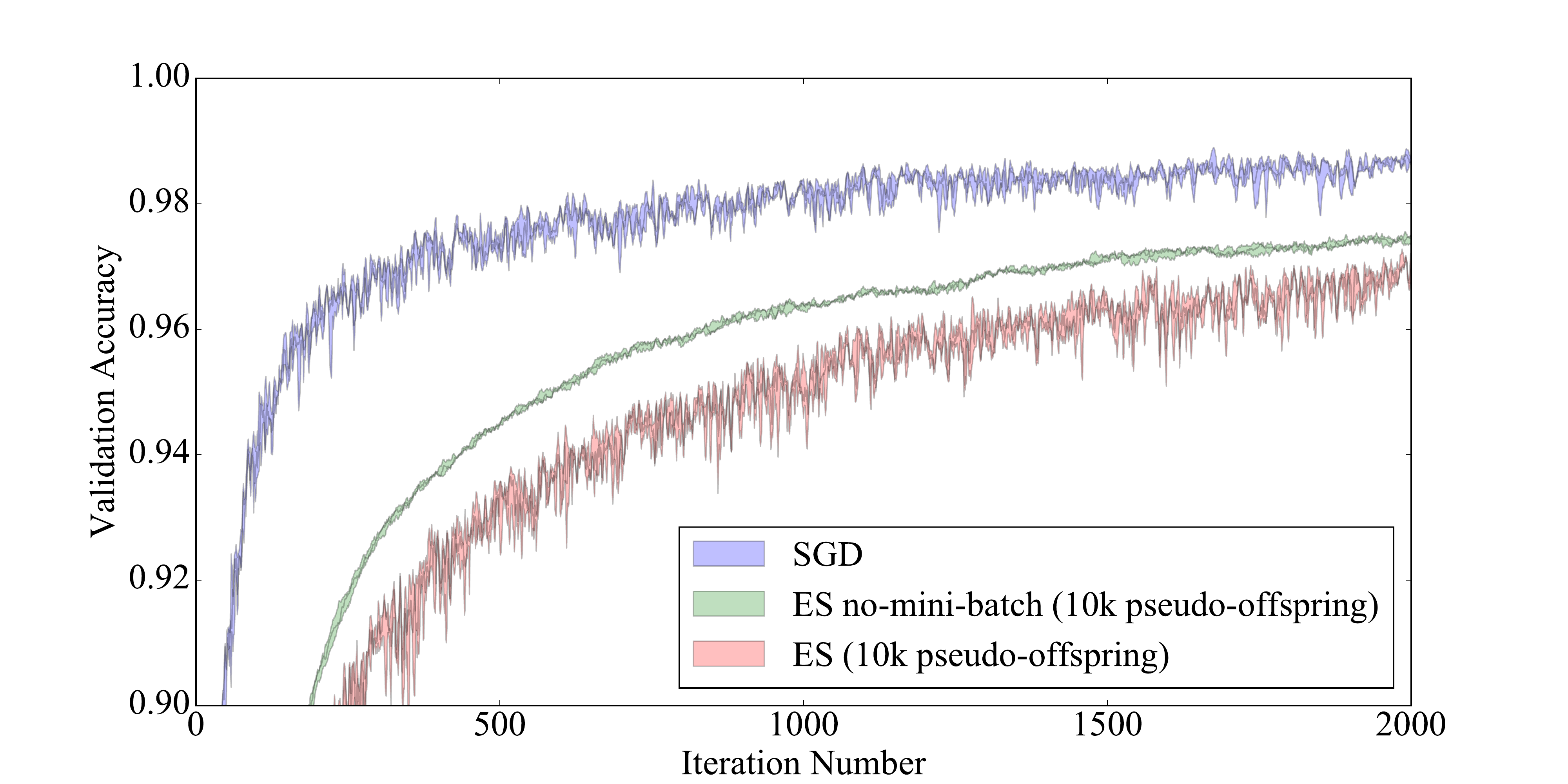}
  \caption{\textbf{Comparing the performance of ES with and without mini-batches.} For each method, the median validation accuracy across 5 runs on MNIST is plotted and the area between the 25th and 75th percentiles is filled. Not only is the learning curve of ES without mini-batches superior to mini-batched ES with the same population size, but it is also much smoother than both mini-batched ES and SGD. The difference in performance for ES without mini-batches compared to with mini-batches is statistically significant with $p < 0.05$ from the Wilcoxon test.
  %\jc{Nice to see the Wicoxon and a p value! But it is odd to mix that with a mean and standard deviation for the reasons stated earlier, so let's switch to median and bootstrapped CI if we can.}\todo{if the p value is lower than lower thresholds, go with that (e.g. 0.01}
  }
  \label{es:no-mini-batch}
\end{figure}

In fact, it turns out that the no-mini-batch ES \emph{does} improve performance. With 10,000 pseudo-offspring over 2,000 iterations, ES without mini-batches achieves an average validation accuracy of 97.44\% across 5 runs  (compared to 96.99\% with mini-batches).
This difference is significant ($p < 0.05$) based on the Wilcoxon test.
The no-mini-batch ES also produces surprisingly smooth learning curves -- in fact much smoother than SGD -- as shown in figure \ref{es:no-mini-batch}.
The smoothness of the curve suggests that removing mini-batches also removes significant noise from the training process, though noise from perturbing $\theta_t$ still remains (but is effectively averaged out to a large extent when many pseudo-offspring are aggregated).

While it is unlikely that ES will ever be a natural choice for supervised learning where perfect gradient information is available, if it were to be applied in practice, it appears that the no-mini-batch variant is the preferred choice.  Unfortunately, when mini-batches are removed it is no longer possible to compute a direct correlation between the gradient from SGD (which learns from mini-batches) and ES (which now does not), but it is nevertheless clear that the no-mini-batch variant is achieving a level of performance that would require a higher level of correlation than that observed in the analogous mini-batched variant.  The intriguing consideration with respect to RL, where gradient information is imperfect and likely noisy, is the potential implication of the much smoother curve that ES seems to obtain.

\subsection{Achieving 99\% in Practice}

To gain further understanding of ES, we are interested in the question of whether ES can achieve near state-of-the-art results for MNIST. In particular, it is interesting to see whether the improved ES without mini-batches can reach 99\% accuracy. Due to limited computational resources and the likely need for many iterations and many pseudo-offspring, we used a smaller network with 28,938 parameters to explore this question. Specifically, the network contains two convolutional layers with 16 and 32 kernels per convolutional layer (both with stride of 5), respectively, followed by ReLU activation functions and 2-by-2 max pooling. The final layer is fully connected with 10 output units. L2 regularization with a coefficient of 0.0002 is applied to mitigate overfitting. It is worthwhile to point out that preliminary results showed that regularization via dropout hurt performance in this particular setup, perhaps because ES itself has enough stochasticity already.  The no-mini-batch approach to ES is followed.

\begin{figure}[h]
  \centering
  \includegraphics[width=\linewidth]{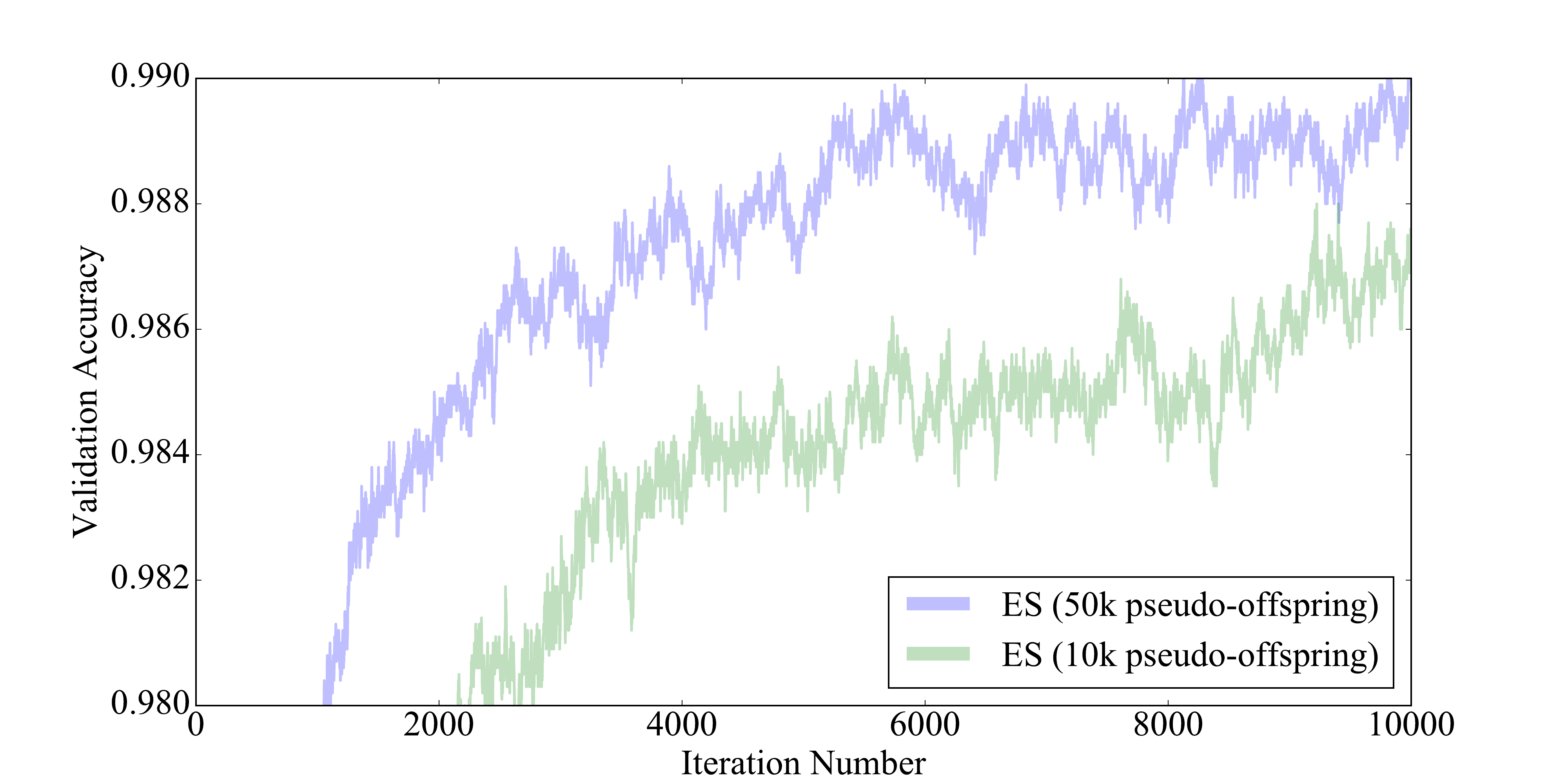}
  \caption{\textbf{99\% validation accuracy in MNIST achieved by ES without mini-batches.} (Note the figure is zoomed significantly compared to figure \ref{es:no-mini-batch}, which is why it looks less smooth.) With 50,000 pseudo-offspring, ES reaches the 99\% threshold, giving a sense of the potential of gradient approximation in ES to make reasonable estimations.  While the added computational expense of ES is unnecessary in the supervised case, this understanding of the approximation potential of ES is useful in the context of RL.  In RL domains, ES gains increasing accuracy in its gradient estimate with increasing parallelism, while also potentially exacting a lower computational cost in doing so.  Thus it may prove under certain conditions a practical choice.}
%  \jc{we need more caveats here to prevent laughter. It's NOT reasonable to spend so much computing to do something one computer can do. So we need to add more words and also acknowledge and defuse that criticism (here and throughout the paper). Here, we can say that of course this computational expense is unnecessary, but it is interesting because of the insight it provides and because in RL domains we do not have true gradients AND thus non-parallelizable gradient estimators are slow, but ES can do great with these sorts of tricks and lots of parallel computing, meaning it may ultimately be the best choice....etc...}.\jc{In all plots with lines for legends, it is virtually impossible in print (and even hard digitally) to tell these pastel colors apart in the legend lines, because they are thin. Can you make them much thicker, or just make a square of that color instead of a line?}}
  \label{99:accuracy}
\end{figure}

As shown in figure \ref{99:accuracy}, ES with 10,000 pseudo-offspring is able to achieve an accuracy around 98.7\%, but it fails to go above the threshold of 99\% even when the number of iterations is doubled to 20,000. However, if the number of pseudo-offspring per iteration is increased to 50,000, ES indeed manages to achieve 99\% accuracy, a first for an evolutionary approach as far as we know.

\section{Discussion}

The main implication is not that ES is appropriate for supervised learning. Usually in supervised learning the optimal gradient (at least per mini-batch) is perfectly accessible, and  even with parallelism the added overhead in ES of tens of thousands or more additional network activations per update of $\theta$ is clearly unwarranted.  However, in RL (as well as other semi-supervised or unsupervised domains) the optimal gradient is not accessible to SGD-based methods either, which makes the results in this paper relevant.  In particular, it is interesting that ES can in principle approximate how SGD would perform in the ideal case with a level of computation that appears realistic today or in the near future.
However, the case for ES in RL encompasses more than matching the gradient of SGD -- in some domains ES can reduce the level of noise further by adding more pseudo-offspring, eventually surpassing the capacity of SGD to point in the most fruitful direction.  With its low-cost parallelism, it can then also gain an advantage in wall-clock time as well.  The results from \citet{es} already provide evidence supporting this perspective.    
%\jc{but that just means they would be tied. what this paper lacks is an argument about why ES for RL will one day become BETTER, which I think is true since SGD and ES are equally hobbled in having a noisy gradient estimate, but ES can potentially reduce the noise further via more pseudo-offspring, and thus once it has a better estimate than SGD methods, it could do better in wall-clock time. But isn't that future already here, as reported in Salimans? If so, we do not need to argue for this happening one day, but instead say why it is already a better choice, no? Also, are there other reasons RL might one day be better?} 
Even if ES falls short of a perfect gradient estimation, the ability to approximate a reasonable gradient may still be sufficient to compete in domains without perfect gradient information available with respect to final performance.

The ability of ES to directly perturb policy parameters means that the gradient estimate is actually approaching the optimal gradient with respect to $\theta_t$ in the limit as the pseudo-offspring size increases.  Recent results such as work from \citep{plappert:arxiv17} on parameter space noise for exploration in RL have also highlighted that directly perturbing policy parameters can offer the benefit of correlated exploration in action space. The gradient approximating ability of ES demonstrated here, as well as results from ES in RL reported by \citet{es}, suggest that the option to explore directly in parameter space through ES taps into similar potential. 
%\jc{that makes it sound like it may happen one day. also, I don't think this is a `may', but is true. ES does get this benefit. Finally, you may wish to look at the GA paper paragraph on this subject for additional or different terms and explanations and cites. We debated it at length and then ran it by Peter. }
Note however that it would be wrong to speculate that one approach (e.g.\ to RL) is necessarily better than the other -- on the contrary, the more nuanced insight is that the breadth of the toolbox of options is widened, and even new hybrid ideas become feasible. 
%It is likely that directly searching in policy space carries its own pros and cons, such as the ability to expose quickly what the most promising proximal policies are in the space around $\theta$.\jc{you just mentioned the deep exploration benefit...so it's a bit odd to not say that here too, and it's also not clear if that is the same thing as this}  This ability could be either a blessing or a curse\jc{as stated, it would always be a blessing. why would that be a curse?}, depending on the context, and PPO \jc{we never said what this is...presumably an edit in the previous paragraph removed that}hints at the potential for synergistic interactions between different kinds of search in RL.

A more subtle, but important point is that while $\sigma$, fixed at 0.002 in the MNIST experiments, is small, in a noisy RL domain $\sigma$ would likely need to be larger to overcome interference from domain noise.  Interestingly, the larger $\sigma$ becomes, the more ES diverges from a traditional finite-difference approximator and begins to optimize not just $\theta$, but the entire distribution of perturbations.  The implications of this property, though not studied in this paper, are studied in a companion paper \citep{lehman:arxiv17fd}, and are likely important for understanding not just the expected performance, but also the expected robustness of RL solutions from ES, and therefore deserve further investigation.  
 
%TODO: if possible and other paper is ready, cite the Joel-Jay paper on ES robustness properties

Of course, it is possible that $\sigma$ can simply be annealed for any domain, which might offer a practical alternative to setting it to a single fixed value.  Or, a more principled approach might be to apply a full-blown natural evolution strategy (NES) \citep{wierstra:jmlr14}, which would encompass optimizing both $\theta$ and $\sigma$.  The extent to which lessons learned about the ES in this paper also extend (or even work better) in the context of NES is an important subject for future study.  Another important caveat is that lessons drawn from MNIST may not extend to other search spaces even of similar dimensionality to those studied here.  Thus the results here should be taken as a hint at the underlying potential of ES rather than an accurate prediction of expected performance in general.

As a practical matter, the diversity of means of applying ES, such as through limited perturbations and with no mini-batches (which, while applied in the supervised case in this paper, may in part reflect a set of noisy rollouts in RL), highlights that ES is indeed a different paradigm from SGD, where such considerations do not exist. One benefit e.g.\ of limited perturbations is simply a significant speedup; especially when perfect accuracy is neither essential nor attainable (which is likely true in many RL domains), such dramatic speedup may begin to eclipse any intrinsic advantage of SGD -- at least in the case of policy gradients, the need to pass gradients among processing units already dampens any such advantage.  For the no-mini-batch approach, the fact that its learning curves are so much smoother than those generated by SGD may have implication in RL, where in effect different rollouts for different pseudo-offspring resemble to some extent a no-mini-batch ES in the supervised case.
However, this analogy is not perfect because rollouts in RL are not sampling from the full breadth of possible experiences.  Thus the broader implications of the smoother learning curve remain speculative pending further investigation.

The more general and perhaps surprising result that an evolutionary algorithm (EA) of any kind can now effectively optimize networks of millions of parameters also raises questions on potentially overlooked capabilities of evolutionary algorithms beyond just ES.  In fact, the sudden realization that evolution can work on deep networks is reminiscent of a similar revelation years ago about SGD and backpropagation \citep{hinton:science06} in DNNs that overturned assumptions on their limits as well.  It is possible that now computation (in the form of massive parallelism) is finally unlocking the potential of evolution as well, opening up new opportunities  that were inconceivable only recently.  
In this spirit, we should not necessarily dismiss the possibility that other EAs in general may begin to yield surprising such results as well.
In fact, in a parallel paper \citep{such:arxiv17}, we show that a simple GA actually can compete with modern RL techniques in some Atari game domains.
While ES undoubtedly follows the gradient more accurately than a traditional EA \citep{dejong:book02}, it is possible that gradient approximation is not always 
the most advantageous strategy for population-based search.
%TODO: connect to Catalyst paper and SMOG paper here?

Finally, it is interesting to consider how SGD and ES might interact fruitfully in the future.  A population of perturbations in effect contains more information than a single suggested gradient direction. Can such a population perhaps be leveraged to complement gradient knowledge? Can these kinds of computations be interleaved?  Can they shed light on each other and perhaps yield a more fundamental understanding overall? This paper provides only an initial glimpse at the potential for understanding the relationship between SGD and ES more deeply.

\section*{Acknowledgements} 
We thank all of the members of Uber AI Labs, in particular Rui Wang, Joel Lehman, and Thomas Miconi for helpful discussions. We also thank Justin Pinkul, Mike Deats, Cody Yancey, Joel Snow, Leon Rosenshein and the entire OpusStack Team inside Uber for providing our computing platform and for technical support.

%!-drawing supervised to RL implications
%   -connection to PPO/other RL literature on parameter perturbations

%!-implications of no mini-batch and limited perturbation results
%   -note smoothness of training curve versus SGD

%!-Jeff: "add to discussion the fact that the future work is need to know to what extent lessons learned on the OpenAI-ES outlined in this paper apply to NES more broadly (including adaptive sigma, and the covariance-NES)"

%!-ES versus GA thoughts

%!-General prospects for ES and evolution as a tool for deep learning with large and deep NNs

%TODO(Ken): possibly delete extra URL and padding from bibtex entries

\bibliographystyle{plainnat}
\bibliography{nnstrings,nn,ucf}

\end{document}